\documentclass{article}
\usepackage{ijcai19}

\usepackage{times}
\usepackage{soul}
\usepackage{url}
\usepackage[hidelinks]{hyperref}
\usepackage[utf8]{inputenc}
\usepackage[small]{caption}
\usepackage{graphicx}
\usepackage{cmath}
\usepackage{amsmath}
\usepackage{booktabs}
\usepackage{algorithm}
\usepackage{algorithmic}
\usepackage{comment}
\usepackage{multirow} 
\usepackage{multicol} 
\usepackage{subcaption}
\usepackage{color}

\newcommand{\xwc}[1]{\textcolor{red}{\bf [Comments: #1] }}

\title{ Knowledge Amalgamation from Heterogeneous Networks\\ 
by Common Feature Learning}

\author{
Sihui Luo$^1$
\and
Xinchao Wang$^2$\and
Gongfan Fang$^1$\and
Yao Hu$^3$\and
Dapeng Tao$^4$\And
Mingli Song$^1$
\affiliations
$^1$Zhejiang University\\
$^2$Stevens Institute of Technology\\
$^3$Alibaba Group\\
$^4$Yunnan University
\emails
\{sihuiluo829, fgfvain97, brooksong\}@zju.edu.cn, 
xinchao.wang@stevens.edu, 
yaoohu@alibaba-inc.com, 
dptao@ynu.edu.cn
}

\begin{document}

\maketitle

\begin{abstract}

An increasing number of well-trained deep networks have been released online by researchers and developers, enabling the community to reuse them in a plug-and-play way without accessing the training annotations. However, due to the large number of network variants, such public-available trained models are often of different architectures, each of which being tailored for a specific task or dataset. In this paper, we study a deep-model reusing task, where we are given as input pre-trained networks of heterogeneous architectures specializing in distinct tasks, as teacher models. We aim to learn a multitalented and light-weight student model that is able to grasp the integrated knowledge from all such heterogeneous-structure teachers, again without accessing any human annotation. To this end, we propose a common feature learning scheme, in which the features of all teachers are transformed into a common space and the student is enforced to imitate them all so as to amalgamate the intact knowledge. We test the proposed approach on a list of benchmarks and demonstrate that the learned student is able to achieve very promising performance, superior to those of the teachers in their specialized tasks.


\end{abstract}

\section{Introduction}\label{sec:1}
\begin{figure*}[t]
  \centering
  \centerline{\includegraphics[width=0.6\linewidth]{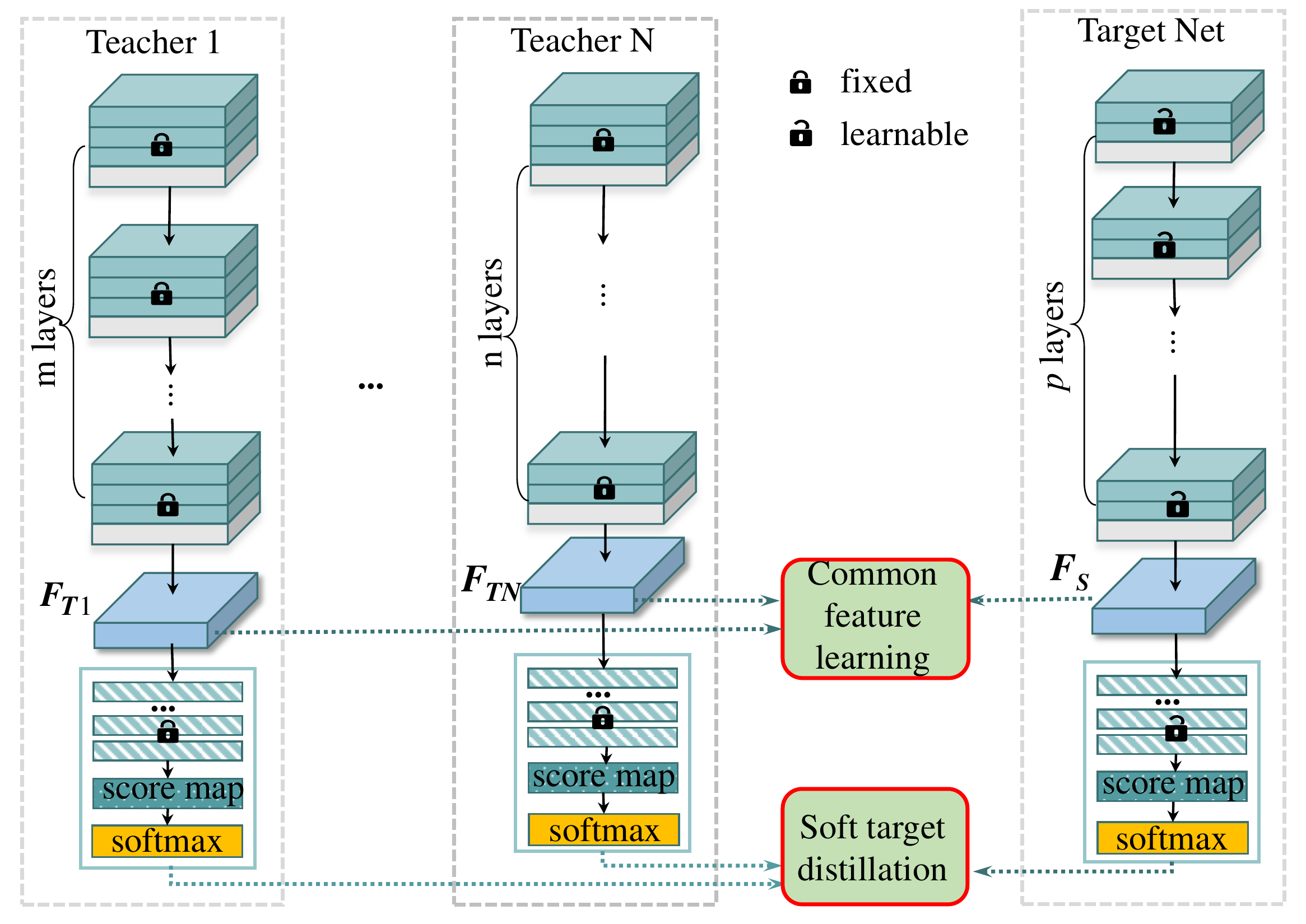}}
\vspace{-1em}
  \caption{\small {Illustration of the proposed heterogeneous knowledge amalgamation approach.  
The student and the teachers may have different
network architectures.
The student learns the transformed features of the teachers
in a common feature space, as depicted by the \emph{common feature learning} block, 
and meanwhile imitates
the soft target predicted by the teachers, as depicted by the \emph{soft target distillation} block.}}
  \label{fig:framework}  
  \vspace{-1.2em}
\end{figure*}

In recent years, deep neural networks~(DNNs) have achieved impressive success in many artificial intelligence tasks like computer vision and natural language processing. Despite the exceptional results achieved, DNNs are known to be data hungry, meaning that very large number of annotations, sometimes confidential and unavailable to the public, are required to train the models. 

To alleviate the reproducing effort, many researchers have released their well-trained models online, 
so that the users may download and use them immediately in a plug-and-play manner. Such publicly-available trained models are, however, often of different architectures, due to rapid development of deep learning and the consequent huge number of network variants, each of which optimized for its own specific tasks or datasets.

In this paper, we investigate an ambitious deep-model reusing task, whose goal is to train a light-weight and multitalented student model, using multiple heterogeneous-architecture teacher models that specialize in different tasks. We assume we have no access to any human annotation but only well-trained teacher models, and focus on how the student model is able to amalgamate knowledge from the heterogeneous teachers.

As we impose no assumption on the teachers' architecture being the same, it is infeasible to conduct a layer-wise knowledge amalgamation. We thus resort to an alternative approach. We project the final heterogeneous features of the teachers into a shared feature space, which is learned, and then enforce the student to imitate the transformed features of the teachers. The student model, learned in this way without accessing human annotations, amalgamates the integrated knowledge from the heterogeneous teachers and is capable of tackling all the teachers' tasks.

The proposed approach is therefore different from and more generic than the conventional knowledge distillation~(KD)~\cite{hinton2015distilling} setup, as a student is supposed to amalgamate knowledge from multiple heterogeneous teachers in the former case yet inherit knowledge from a single teacher in the latter.
It also differs from Taskonomy~\cite{Zamir2018Task}, 
which aims to solve multiple tasks in one system with fewer annotated data by finding transfer learning dependencies.

Specifically, we focus on knowledge amalgamation on the classification task, and test the proposed approach on a list of benchmarks of different themes. Results show that the learned student, which comes in a compact size, not only handles the complete set of tasks from all the teachers but also 
achieves results superior to those of the teachers in their own specialities. 

Our contribution is thus a knowledge amalgamation approach tailored for 
training a student model, without human annotations, using heterogeneous-architecture teacher models that specialize in different tasks. This is achieved by learning a common feature space, wherein the student model imitates the transformed features of the teachers to aggregate their knowledge. Experimental results on a list of classification datasets demonstrate the learned student outperforms the teachers in their corresponding specialities.

\section{Related Work}\label{sec:2}

\paragraph{Model ensemble.} The ensemble approach combines predictions from a collection of models
by weighted averaging or voting~\cite{Hansen1990neural,Dietterich2000Ensemble,wang2011tip}. It has been long observed that ensembles of multiple networks are generally much more robust and accurate than a single network. Essentially, ``implicit'' model ensemble usually has high efficiency during both training and testing. The typical ensemble methods include: Dropout~\cite{Srivastava2014DropoutAS}, Drop Connection~\cite{wan2013reg}, Stochastic Depth ~\cite{huang2016deep}, Swapout~\cite{Singh2016Swap}, etc. These methods generally create an exponential number of networks with shared weights during training and then implicitly ensemble them at test time. 
Unlike model ensemble, our task here aims to train a single student model that 
performs the whole collection of tasks handled by all teachers, with no human annotations.




\paragraph{Knowledge distillation.} The concept of knowledge distillation is originally proposed by Hinton~et~al~\cite{hinton2015distilling} for model compression~\cite{WangCVPR18}.
It uses the soft targets generated by a bigger and deeper network to train a smaller and shallower network and achieves similar performance as the deeper network. Some researches~\cite{Romero2015FitNets,wang2018progressive} extended Hinton's work~\cite{hinton2015distilling} by using not only the outputs but also the intermediate representations learned by the teacher as hints to train the student network. Most knowledge distillation methods~\cite{hinton2015distilling,Romero2015FitNets,wang2018progressive} fall into single-teacher single-student manner, where the task of the teacher and the student is the same. Recently, Shen~et~al~\cite{cs-aaai2019} proposes to reuse multiple pre-trained classifiers, which focus on different classification problems, to learn a student model that handles the comprehensive task. Despite the superior effectiveness of their N-to-1 amalgamation method, they impose a strong assumption that the network architectures of the teacher are \emph{identical}.
The work of~Ye~et~al~\cite{YeCVPR19}, similarly, imposes the same assumption on the network architectures. Our method, by contrast, has no such constraints and is capable of resolving the amalgamation of heterogeneous teachers.


\paragraph{Domain adaption.} Domain adaptation~\cite{ben2010theory,GongICML16}, which belongs to transfer learning,  aims at improving the testing performance on an unlabeled target domain while the model is trained on a related yet different source domain. Since there is no label available on the target domain, the core of domain adaptation is to measure and reduce the discrepancy between the distributions of these two domains. In the literature, Maximum Mean Discrepancy (MMD)~\cite{gretton2012kernel} is a widely used criterion for measure the mismatch in different domains, which compares distributions in the Reproducing Kernel Hilbert Space (RKHS). We extend MMD to match the distribution of the common transferred features of the teachers and the student in this paper such that our method can be more robust to amalgamation of heterogeneous teachers which may be trained on datasets across domains.

\section{Overview}\label{sec:3}
We assume we are given $N$ teacher networks, each of which denoted by 
$Ti$. Our goal is to learn a single student model $S$ that amalgamates the knowledge of the teachers and masters the tasks of them, with no annotated training data.
The architectures of the teacher models can be identical or different, with {no specific constraints}.  

The overall framework of the proposed method is shown in Fig.~\ref{fig:framework}. 
We extract the final features from the heterogeneous teachers and then project them into a learned common feature space, in which the student is expected to imitate the teachers' features, as depicted by the \emph{common feature learning} block.
We also encourage the student to produce the same prediction as the teachers do, depicted by the \emph{soft target distillation} block.
In next section, we give more details on these blocks and our complete model. 


\section{Knowledge Amalgamation by Common Feature Learning}\label{sec:4}

In this section, we give details of the proposed knowledge amalgamation approach. 
As shown in Fig.~\ref{fig:framework}, the amalgamation consists of two parts: feature learning in the common space, represented by the \emph{common feature learning} block, and classification score learning, represented by the \emph{soft target distillations} block. In what follows, we discuss the two blocks and then give our final loss function.

\begin{figure}[t]
  \centering
  \centerline{\includegraphics[width=0.85\linewidth]{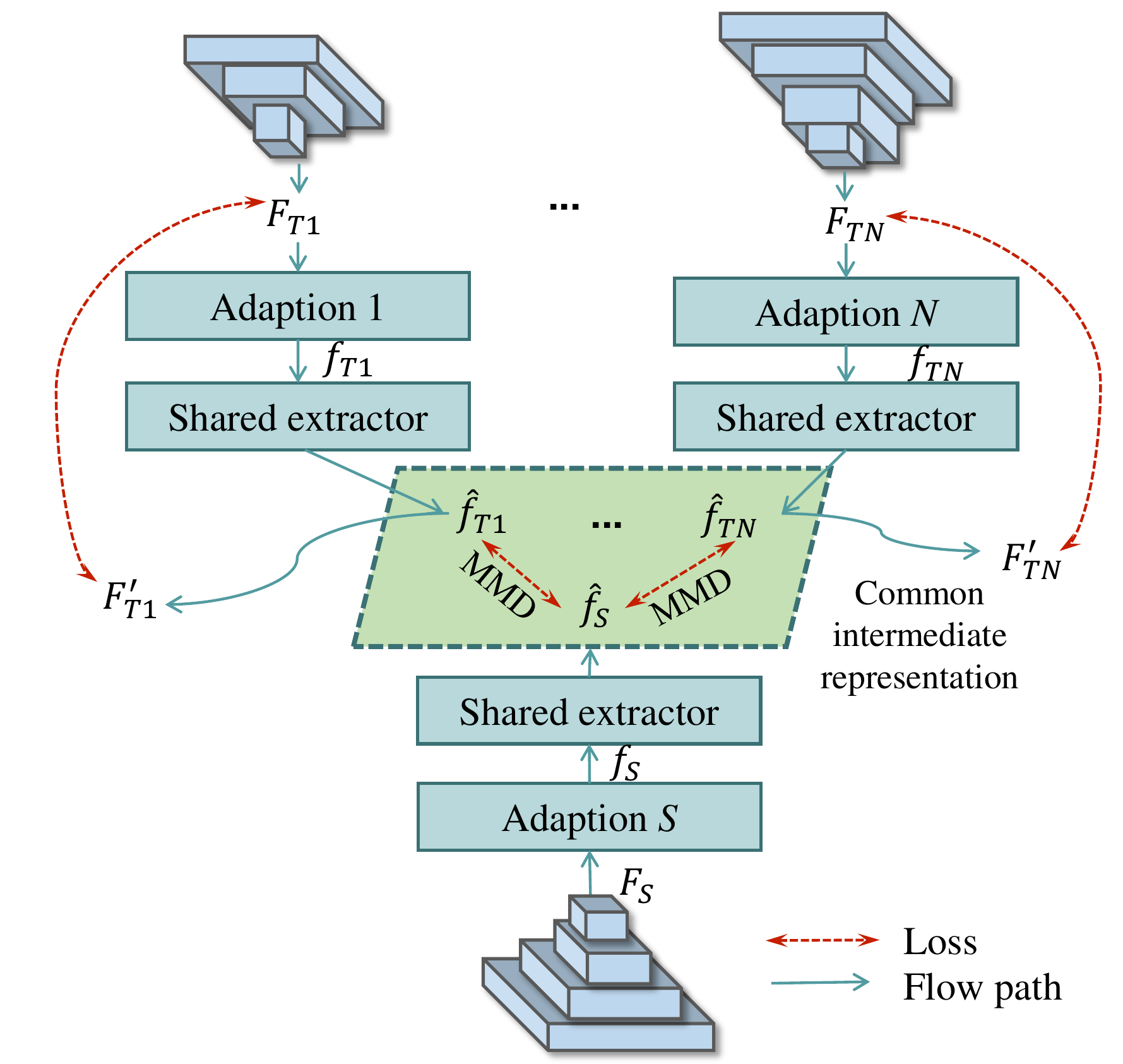}}
  \caption{\small {Illustration of the \emph{common feature learning} block.
  Two types of losses are imposed: the first on the distances between the transformed features of the student~(target net) and those of the teachers in the common space, and second on the reconstruction errors of the teachers' features mapping back to the original space.
  }}
  \label{fig:comFeature}  
\end{figure}
\subsection{Common Feature Learning } \label{sec:main_method}
The structure of the common feature learning block is shown in Fig.~\ref{fig:comFeature}. 
The features of the teachers and those to be learned of the students 
are transformed to a common features space, via an adaption layer and
a shared extractor, for which both the  parameters are learned.
Two loss terms are imposed: the feature ensemble loss $\mathcal{L}_{M}$ 
and the reconstruction loss $\mathcal{L}_{R}$. The former one encourages 
the features of the student to approximate those of the teachers in the common space,
while the latter one ensures the transformed features can be mapped back 
to the original space with minimal possible errors.


\subsubsection{Adaption Layer}
The adaption layer is used to align the output feature dimensions of the teachers and that of the student to be the same. Thus, the teachers and the student have their individually-parameterized adaption layers. The layers are formed by convolutions with 1$\times$1 kernel~\cite{szegedy2015going} that generate a predefined length of output with different input sizes. 
We use $F_S$ and $F_{Ti}$ to denote respectively  
the features of the student and teacher $Ti$ before being fed to the adaption layer,
and use $f_S$ and $f_{Ti}$ to denote respectively their aligned features after passing through the adaption layer. The channel number of $f_S$ and $f_{Ti}$ is 256 in our implementation.


\subsubsection{Shared Extractor}
Once the aligned output features are derived, a straightforward approach would be to directly average the features of the teachers $f_{Ti}$ as that of the student $f_S$. 
However, due to domain discrepancy of the training data and architectural difference of the teacher networks, the roughly aligned features may still remain heterogeneous. 


To this end, we introduce a small and learnable sub-network, of which the parameters are shared among the teachers and the student as shown in Fig.~\ref{fig:comFeature}, 
to convert all features to a common and interactable space, which we call the \emph{common feature space}.
This sub-network, termed shared extractor, 
is a small convolution network formed by three residual block of 1$\times$1 stride. 
It converts $f_{Ti}$ and $f_s$ to 
the transformed features $\hat{f}_{Ti}$ and $\hat{f}_{S}$ in the common space,
of which the channel number is taken to be 128 in our implementation.



\subsubsection{Knowledge Amalgamation}
To amalgamate knowledge from the heterogeneous teachers, we enforce the features of the student to imitate those of the teachers in the shared feature space. 
Specially, we adopt the Maximum Mean Discrepancy (MMD)~\cite{gretton2012kernel}, traditionally used to measure the domain mismatch in domain adaption task, in aim to align the student and teacher domains via estimated posteriors. MMD can be regarded as a distance metric for probability distributions~\cite{gretton2012kernel}. 
In our case, we measure the discrepancy between the 
output features of the student and those of the teachers.

We first focus on the discrepancy between the student and one teacher. 
We use $\mathcal{X}=\{ \hat{f}_{T}^i\}_{i=0}^{C_T}$ to denote the 
set of all features of the teacher, where ${C_T}$ is the total number of the teacher's features.
Similar, we use $\mathcal{Y}=\{ \hat{f}_{S}^i\}_{i=0}^{C_S}$ to denote 
all features of the student, where ${C_S}$ is the total number of the student's features.
An empirical approximation to the MMD distance of $\hat{f}_T$ and $\hat{f}_S$ is computed as follows:
\begin{equation}\label{equ:1}
\mathcal{MMD}=\Vert \frac{1}{C_T} \sum_{i=1}^{C_T}\phi(f_{T}^i) - \frac{1}{C_S} \sum_{j=1}^{C_S}\phi(f_{S}^j) \Vert ^2_2,
\end{equation}\noindent
where $\phi$ is an explicit mapping function. By further expanding it using the kernel function
, the MMD loss is given as:
\begin{align}\label{equ:2}
\ell_{m} (\hat{f}_T,\hat{f}_S)= &\sum_{i=1}^{C_T} \sum_{j=1}^{C_T}{ \frac{ \mathcal{K}(\hat{f}_T^i,\hat{f}_T^j)}{C_T^2} } - \sum_{i=1}^{C_T} \sum_{j=1}^{C_S}{ \frac{2 \mathcal{K}(\hat{f}_T^i,\hat{f}_S^j)}{C_S C_T}} +\nonumber\\
&+\sum_{i=1}^{C_S} \sum_{j=1}^{C_S}{ \frac{\mathcal{K}(\hat{f}_S^i,\hat{f}_S^j)}{C_S^2} },
\end{align}\noindent
where $\mathcal{K}(\cdot, \cdot)$ is a kernel function that projects sample vectors into a higher dimensional feature space. We note that we calculated $\ell_{m}$ with normalized $\hat{f}_T^j$ and $\hat{f}_S^j$. 


We then aggregate all such pairwise MMD losses between each teacher and the student, as shown in 
Fig.~\ref{fig:comFeature}, and write the overall discrepancy $\mathcal{L}_{M}$ in the common space as
\begin{equation}\label{equ:3}
\mathcal{L}_{M} = \sum_{i=1}^{N}{ \ell_{m} (\hat{f}_{Ti}, \hat{f}_S)},
\end{equation}
where $N$ is the number of teachers.

To further make the learning of common feature space more robust, we add a auxiliary reconstruction loss to make sure that the transferred feature can be ``mapped back''. Specially, we reconstruct the original output feature of the teacher using the transferred ones. Let $F_{Ti}^\prime $ denote the reconstructed feature of teacher $Ti$, the reconstruction loss $\mathcal{L}_{R}$ is defined as
\begin{equation}\label{equ:4}
\mathcal{L}_{R} =\sum_{i=1}^{N} \| F_{Ti}^\prime - F_{Ti}\|_2,
\end{equation}

\subsection{Soft Target Distillation}
Apart from learning the teachers' features, the student is also expected to produce identical or similar predictions, in our case classification scores, as the teachers do. We thus also take the predictions of the teachers, by feeding unlabelled input samples, and as supervisions for training the student.

For disjoint teacher models that handle non-overlapping classes, we directly stack their score vectors, 
and use the concatenated score vector as the target for the student. The same strategy can in fact also be used for teachers with overlapping classes: at training time, we treat the  multiple entries of the overlapping categories as different classes, yet during testing, take them to be the same category.

Let $c_i$ denote the function of teachers that maps the last-conv feature to the score map and $c_s$ denote the corresponding function of the student.
The soft-target loss term $L_{C}$, which makes the response scores of the student network to approximate the predictions of the teacher, is taken as:
\begin{equation}\label{equ:5}
\mathcal{L}_{C} =\|c_s \circ F_s- [c_1 \circ F_{T1}, \dots, c_N \circ F_{TN} ]\| _2.
\end{equation}

\subsection{Final Loss}
We incorporate the loss terms in Eqs.~(\ref{equ:3}),~(\ref{equ:4}) and~ (\ref{equ:5}) into our final loss function. The whole framework is trained end-to-end by  optimizing the following objective:
\begin{equation}\label{equ:6}
\mathcal{L} = \alpha\mathcal{L}_{C}+(1-\alpha)( \mathcal{L}_{M}+\mathcal{L}_{R}), \alpha \in [0,1]
\end{equation}\noindent
where $\alpha$ is a hyper-parameter to balance the three terms of the loss function. By optimizing this loss function, the student network is trained from the amalgamation of its teacher without annotations.

\section{Experiment}\label{sec:5}
We demonstrate the effectiveness of the proposed method on a list of benchmark, and provide implementation details, comparative results and discussions, as follows.

\begin{table}[tp]
\small
\begin{center}
\begin{tabular}{l|r|r|r}
\toprule
{\bf Classification Dataset} & {\bf Categories } & {\bf Images } & {\bf Train/Test}\\
\midrule
Stanford Dog     &  120  &  25,580 & Train/Test \\ 
Stanford Car     &  195  &  16,185 & Train/Test \\
CUB200-2011      &  200  &  11,788 & Train/Test \\
FGVC-Aircraft    &  102  &  10,200 & Train/Test \\
Catech 101       &  101  &  9,146  & Train/Test \\
\midrule
{\bf Face Dataset} & {\bf Identities } & {\bf Images} &{\bf Train/Test}\\
\midrule
CASIA      &10,575 & 494,414 & Train\\
MS-Celeb-1M &100K   & 10M  & Train\\
CFP-FP      &500    & 7000 & Test\\
LFW         &1,680  & 13,000& Test\\
AgeDB-30    &568    &16,488 & Test\\
\bottomrule
\end{tabular}
\end{center}
\vspace{-3mm}
\caption{Datasets for training and testing. Train/Test: usage for training or testing}
\label{tab: dataset}
\end{table}

\subsection{Networks}
In this paper, we utilize the widely used $alexnet$, $vgg$-16, and the $resnet$~\cite{He2016Deep} family including $resnet$-18, $resnet$-34, and $resnet$-50,
as our model samples. 
Also, we use the baseline Arcface~\cite{deng2018arcface} network with basic cross entropy loss
for the face recognition task.


\subsection{Dataset and Training Details}
We test the proposed method on a list of classification datasets summarized in Tab.~\ref{tab: dataset}. 

Given a dataset, we pre-trained the teacher network against the one-hot image-level labels in advance over the dataset using the cross-entropy loss. We randomly split the categories into \emph{non-overlapping}
parts of equal size to train the  heterogeneous-architecture teachers.
The student network $S$ is trained over non-annotated training data with the score vector and prediction of the teacher networks. Once the teacher network is trained, we freeze its parameters when training the student network.


In face recognition case, we employ CASIA webface~\cite{yi2014learning} or MS-Celeb-1M as the training data,  1,000 identities from its overall 10K ones are used for training the teacher model. During training, we explore  face verification datasets including LFW~\cite{huang:inria-00321923}, CFP-FP~\cite{cfp_fp2016}, and AgeDB-30~\cite{Moschoglou2017AgeDB} as the validation set. 

\subsection{Implement Details}
We implement our method using PyTorch~\cite{He2016Deep} on a Quadro P5000 16G GPU. An Adam~\cite{kingma2014adam} optimizer is utilized to train the student network. The learning rate is $e^{-4}$, while the batch size is 128 on classification datasets
and 64 on face recognition ones.

\subsection{Results and Analysis}

We perform various experiments on different dataset to validate the proposed method. Their results and some brief analysis are presented in this subsection. 

\begin{table}[tp]
\small
\begin{center}
\begin{tabular}{l|r|r}
\toprule
{\bf Model } & {\bf Dog }(\%) & {\bf Catech 101}(\%)\\
\midrule
\textit{\bf T1} ($resnet$-18)     &  45.09  &  71.69   \\ 
\textit{\bf T2} ($resnet$-34)     &  50.65  &  75.36    \\
\midrule
\textbf{ensemble}                     &  37.90  &  65.99 \\
{$resnet$-34} {\bf with GT}      &  53.03  &  72.53     \\
\textbf{KD} ($resnet$-34)         &  48.31  &  73.15  \\ 
\textbf{Ours} ($resnet$-34)      &  \textbf{53.80}  & \textbf{73.69} \\
\bottomrule
\end{tabular}
\end{center}
\vspace{-3mm}
\caption{Comparison of proposed knowledge amalgamation method against others on Stanford dog and Catech 101 dataset, in terms of classification accuracy. 
Given two teachers trained on two class-exclusive splits of the dataset from scratch, 
our method outperforms not only the two teachers on their own specialized domains,
but also model ensemble, the baseline KD, and even training a $resnet$-34 using ground truth data.}
\label{tab: cls1}
\end{table}

    
\subsubsection{Comparison with Other Training Strategy}

\begin{itemize}
    \item \textbf{Model ensemble.} The score vectors of the teachers are concatenated and the input sample is classified to the class with the highest score in the concatenated vector.

    \item \textbf{Training with GT.} We train the student network from scratch with ground-truth annotations. This result is used as an reference of our unsupervised method.
    
    \item \textbf{KD baseline.} We apply the knowledge distillation method~\cite{hinton2015distilling} to heterogeneous teachers extension. The score maps of the teachers are stacked and used to learn the target network.
    
    \item \textbf{Homogeneous Knowledge Amalgamation.} We also compare our results to the state-of-the-art homogeneous knowledge amalgamation method of Shen's~\cite{cs-aaai2019}, when applicable, since this method works only when the teachers share the same network architecture. 
    
\end{itemize}

\begin{table}[!t]
\small
\begin{center}
\begin{tabular}{c|rrrr}
\toprule
{\bf Model}  & {\bf LFW}~(\%)& {\bf Agedb30}~(\%)& {\bf CFP-FP}~(\%) \\
\midrule
$T1$           & 97.43  &  84.72   & 86.20 \\ 
$T2$           & 97.80  &  85.87   & 87.27 \\
\midrule
KD     & 97.15  & 84.97    & 86.87 \\
Ours           & \textbf{98.10}    & \textbf{86.93}   & \textbf{87.73} \\ 
\bottomrule
\end{tabular}
\end{center}
\vspace{-3mm}
\caption{Comparison of the proposed knowledge amalgamation method against the teachers and 
KD on face recognition task. Each teacher is trained with a 3000-classes split from CASIA.}
\label{tab: face1}
\end{table}

We compare the performances of the first three methods and ours in Tab.~\ref{tab: cls1} and Tab.~\ref{tab: face1}, since we focus here on heterogeneous teachers and thus cannot apply Shen's method. As can be seen, the proposed methods, denoted by ours, achieves the highest accuracies. The trained student outperforms not only the two teachers, but also the $resnet$ model that trains with ground truth. The encouraging results indicate that the student indeed amalgamates the knowledge of the teachers in different domains, which potentially complement each other and eventually benefit the student's classification. 


To have a fair comparison with Shen's method that can only handle homogeneous teachers, we adopt the same network architecture, $alexnet$, as done in~\cite{cs-aaai2019}.
The results are demonstrated in Tab.~\ref{tab: shen's}.
It can be observed that our method yields an overall similar yet sometimes much better results (e.g. the case of car).


\begin{table}[t]
\small
\begin{center}
\begin{tabular}{l|cccc}
\toprule
{\bf Model}  & \textbf{Dog} & \textbf{CUB} & \textbf{Car} & \textbf{Aircraft} \\
\midrule
ensemble     & 43.5 & 41.4 & 37.8  & 47.1 \\
KD               & 10.4 & 30.0 & 17.0 & 39.9 \\
\cite{cs-aaai2019}     & \textbf{45.3} & 42.6 & 40.6  & \textbf{49.4} \\
Ours           & 44.4 & \textbf{43.6} & \textbf{45.3} & 47.5\\ 
\bottomrule
\end{tabular}
\end{center}
\vspace{-3mm}
\caption{Comparing our results with homogeneous amalgamation methods on several classification datasets. Note that Dog stands for Stanford Dog, CUB for CUB200-2011, Car for Stanford Car, and Aircraft for FGVC-Aircraft.}
\label{tab: shen's}
\end{table}

\subsubsection{More Results on Heterogeneous Teacher Architectures}
As our model allows for teachers with different architectures, we test more combinations of architectures and show our results Tab.~\ref{tab: archtecture1}, where we also report the corresponding performances of KD. 
We test combinations of teachers with different $resnet$ and $vgg$ architectures.
As can be seen, the proposed method consistently outperforms KD under all the settings. 
Besides, we observe that the student with more complex network architecture~(deeper and wider) trends to gain significantly better performance.

\begin{table}[!t]  
\small
\centering
\begin{tabular}{c|c|rrrr}
\toprule
\multirow{2}*{\bf Teachers} & \multirow{2}*{\bf Method} & \multicolumn{4}{c}{\bf Target Net (acc.\%)}  \\ 
\cline{3-6}
& &  $res$-18 & $res$-34 & $res$-50 &$vgg$-16\\
\midrule
T1 ($vgg$-16) & KD   & 66.4 & 68.6 & 70.5 & 65.6\\
T2 ($res$-18) & Ours & \textbf{66.7} & \textbf{69.1} & \textbf{71.6} & \textbf{66.3}\\
\midrule
T1 ($vgg$-16) & KD   & 64.6 & 68.7 & 70.5 & 65.8\\
T2 ($res$-34) & Ours & \textbf{66.4} & \textbf{69.3} & \textbf{71.2} & \textbf{66.0}\\
\midrule
T1 ($res$-18) & KD   & 76.2 & 80.2 & 82.0 & 79.2 \\
T2 ($res$-34) & Ours & \textbf{77.9} & \textbf{81.3} & \textbf{82.3} & \textbf{79.3}\\
\midrule
T1 ($res$-50) & KD   & 77.6 & 80.4 & 82.5 & 80.2\\
T2 ($res$-34) & Ours & \textbf{78.5} & \textbf{81.6} & \textbf{83.8} & \textbf{80.8}\\
\bottomrule
\end{tabular}
\vspace{-2mm}
\caption{Comparing the results of ours and KD, using teachers of heterogeneous architectures. 
Results are obtained on Stanford Dog, where all the classes are split 
into two non-overlapping parts, each of which is used for training one teacher. Here the teachers are trained with pre-trained models on ImageNet.
}
\label{tab: archtecture1}
\end{table}

\subsubsection{Amalgamation with Multiple Teachers} 
We also test the performance of our method in multiple-teachers case. 
We use a target network of the $resnet$-50 backbone to amalgamate knowledge from two to four teachers, each specializes in a different classification task on different datasets.
Our results are shown in Tab.~\ref{tab: split_num1}. 
Again, the student is able to achieves results that are in most cases superior to those of the teachers. 


\begin{table}[t]  
\small
\centering
\label{tab:methodcompare}
\begin{tabular}{|c|c|c|c|c|}
\hline
\multirow{2}*{\bf Teachers} & \multicolumn{4}{c|}{\textbf{Datasets}~(acc. \%)}  \\ 
\cline{2-5}
            & \textbf{Dog}  & \textbf{CUB} & \textbf{Aircraft} & \textbf{ Car}\\
\hline 
$Ti$ arch  & $resnet$-50 & $resnet$-18 & $resnet$-18 & $resnet$-34 \\
$Ti$ acc   & \textbf{87.1} & 75.6 & 73.2 & 82.9 \\
\hline
2 teachers & 84.3  & \textbf{78.9} & - & -   \\
\hline
3 teachers & 83.1  & 77.7 & \textbf{79.0} & -   \\
\hline
4 teachers & 82.5  & 77.5  & 78.3 & \textbf{84.2}  \\
\hline
\end{tabular}
\vspace{-2mm}
\caption{Comparative results of the teachers and the student trained with varying number of teachers.
The teachers are trained for different classification tasks on distinct datasets.} 
\label{tab: split_num1}
\end{table}

\subsubsection{Supervision Robustness}
Here, we train the target network on two different types of data: 1) the same set of training images as the those of the teachers, which we call \emph{soft unlabeled} data; 2) the totally unlabeled data that are different yet similar with the teacher's training data, which we call  \emph{hard unlabeled} data. The target network is trained with the score maps and the predictions of its teachers as soft targets on both types of training data. 

Specifically, given two teacher models trained on different splits of CASIA dataset, one student is trained using CASIA and therefore counted as soft unlabeled data, and the other student is trained using a totally different source of data, MS-Celeb-1M, and thus counted hard unlabeled data.

We show the results of the student trained with both the types of data 
in Tab.~\ref{tab:plain}.
As expected, the results using soft unlabeled data are better than those using hard unlabeled, but only very slightly.
In fact, even using hard unlabeled data, our method outperforms KD, confirming the robustness of the proposed knowledge amalgamation. 


\begin{table}[t]
\small
\centering
\begin{tabular}{|l|ccc|}
\toprule
{\bf Model}       & \textbf{CFP-FP}(\%)  & \textbf{LFW}(\%) & \textbf{Agedb30}(\%)\\
\midrule
KD            & 86.87  & 97.15   & 86.87 \\
soft unlabeled   & 87.73  & 98.12   & 87.73 \\
hard unlabeled   & 86.36  & 98.12   & 87.00 \\
\bottomrule
\end{tabular}
\vspace{-2mm}
\caption{Comparative results of our student when trained with different types of non-annotated data, under the supervision of teachers on face recognition datasets.}
\label{tab:plain}
\end{table}

\subsubsection{Ablation Studies}
We conduct ablation studies to investigate the contribution of the shared extractor and that of MMD described in Sec.~\ref{sec:main_method}. For shared extractor, we compare the performances by turning it on and off. For MMD, on the other hand, we replace the module with an autoencoder, which concatenates the extracted features of the teachers and encodes its channels to its half size as the target feature for the student network. We summarize the comparative results in Tab.~\ref{tab: ablation}, where we observe that the combination with shared extractor turned on and MMD yields the overall best performance.

\begin{table}[t]
\small
\begin{center}
\begin{tabular}{|c|c|c|c|c|}
\hline
{\bf Shared Ext.} & {\bf AE } & {\bf MMD}  & {\bf Dog}~(\%) & {\bf CUB}~(\%)\\
\hline
\multicolumn{3}{|c|}{KD  } &  80.22 & 75.49\\
\hline
                 &$\surd$   &       & 79.18 & 75.44\\ 
\hline
                 &   &  $\surd$     & 78.75 & 74.25\\ 
\hline
$\surd$    &$\surd$   &    &  78.92 & 74.44\\
\hline
$\surd$    &     & $\surd$  &\textbf{81.34} & \textbf{76.34}\\
\hline
\end{tabular}
\end{center}
\vspace{-3mm}
\caption{Results of the proposed method with 1) shared extractor turned on and off, and 2) the MMD loss replaced by an autoencoder~(AE). The best result is achieved when shared extractor is turned on and MMD loss is utilized.}

\label{tab: ablation}
\end{table}

\subsection{Visualization of the Learned Features}
\begin{figure}[!t]
\begin{minipage}[b]{.99\linewidth}
  \centering
  \centerline{\includegraphics[width=0.85\linewidth]{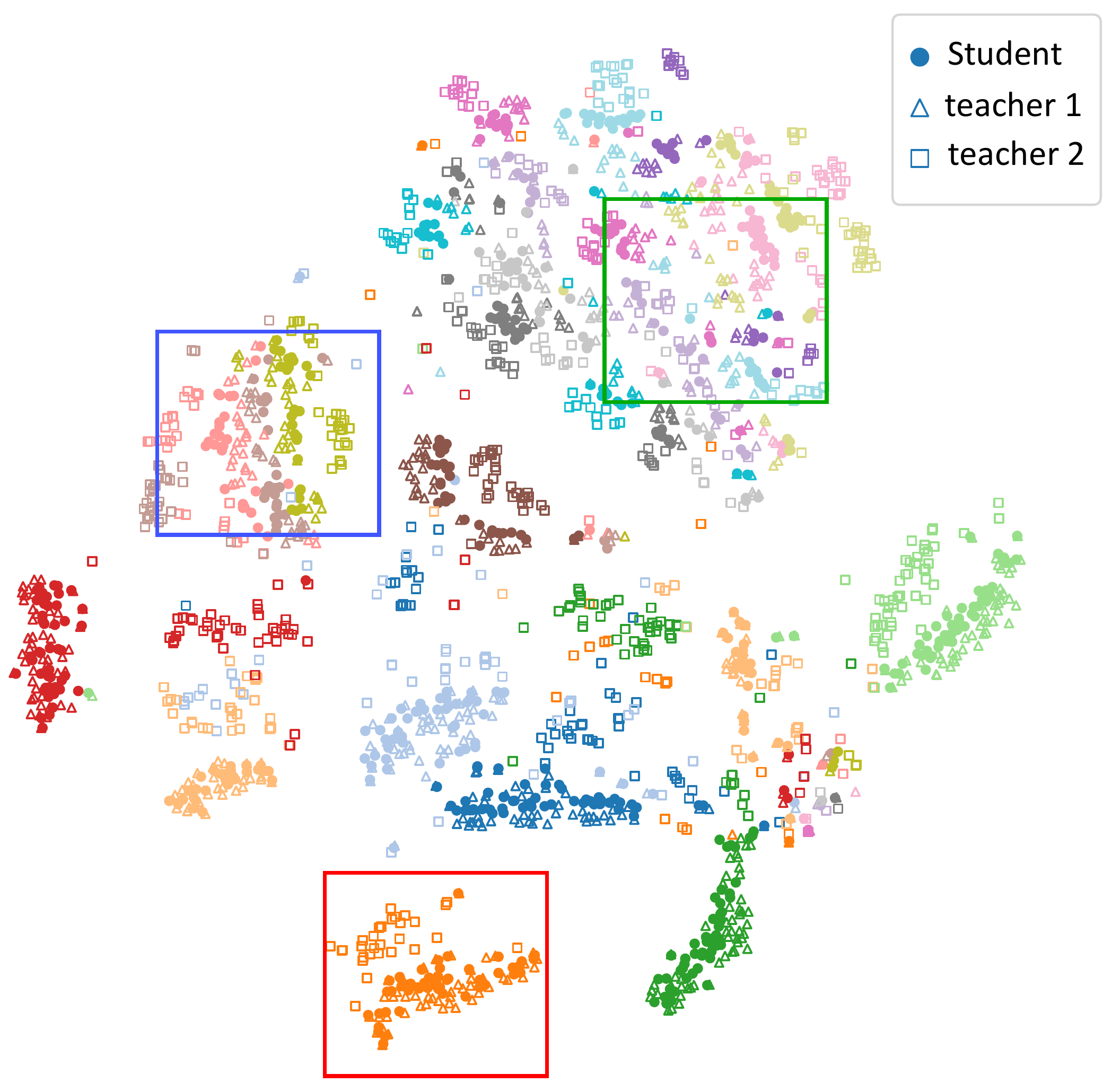}}
  \centerline{(a) t-SNE of $\hat{f}_{Ti}$ and $\hat{f}_{s}$.}\medskip
\end{minipage}
\hfill
\begin{minipage}[b]{0.96\linewidth}
  \centering
  \centerline{\includegraphics[width=0.85\linewidth]{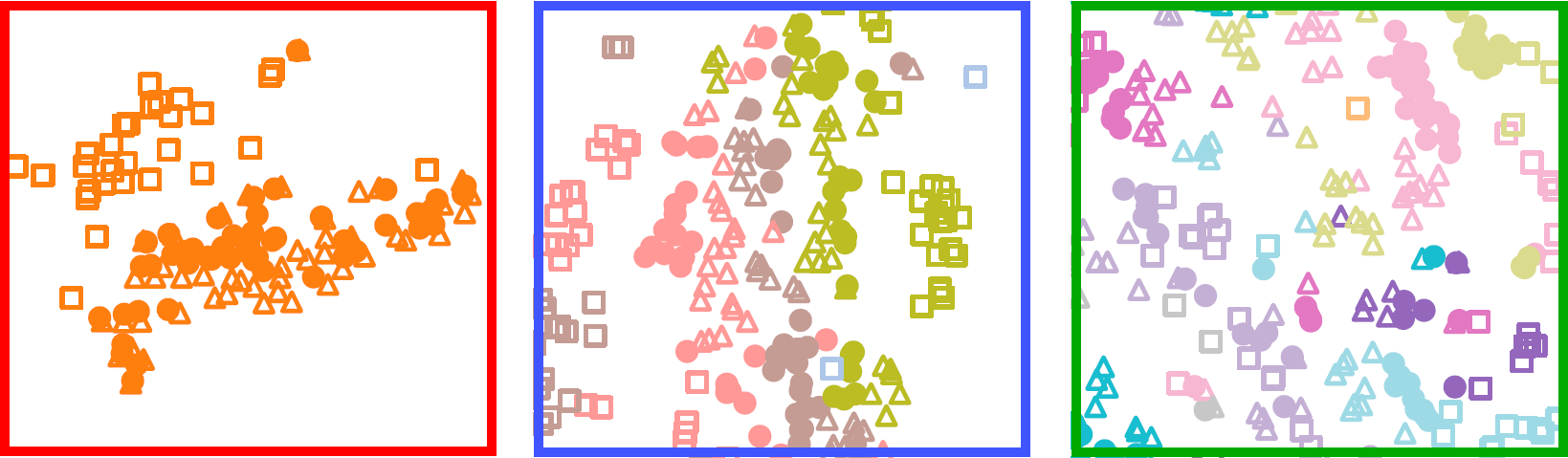}}
  \centerline{(b) Three zoomed-in regions.}\medskip
\end{minipage}
\vspace{-1em}
   \caption{ Visualization of the features of the teachers and those learned by the student in the common feature space. The teachers are trained on Stanford dog and CUB200-2011. We randomly choose test samples of 20 classes to generate this t-SNE plot, where each color denotes a category.} 
\label{fig:visual}
\vspace{-3mm}
\end{figure}

We also visualize in Fig.~\ref{fig:visual} the features of the teachers and those learned by the student in the common feature space. The plot is obtained by t-SNE with $\hat{f}_{Ti}$ and $\hat{f}_{S}$ of 128 dimensions. We randomly select samples of 20 classes on the two-teacher amalgamation case on Stanford dog and CUB200-2011 dataset. We zoom in three local regions for better visualization, where the first one illustrates one single class while last two illustrate multiple cluttered classes and thus are more challenging.
As we can observe, the learned features of the student, in most cases of the three zoomed regions, 
gather around the centers of the teachers' features, indicating the proposed approach is indeed able to amalgamate knowledge from heterogeneous teachers. 

\section{Conclusion}\label{sec:6}
We present a knowledge amalgamation approach for training a student that masters the complete set of expertise of multiple teachers, without human annotations.These teachers handle distinct tasks and are not restricted to share the same network architecture. The knowledge amalgamation is achieved by learning a common feature space, in which the student is encouraged to imitate the teachers' features. We apply the proposed approach on several classification benchmarks, and observe that the derived student model, with a moderately compact size, achieves performance even superior to the ones of the teachers on their own task domain.

\section*{Acknowledgments}
This work is supported by National Natural Science Foundation of China (61572428, U1509206), Fundamental Research Funds for the Central Universities, Stevens Institute of Technology Startup Funding, and Key Research and Development Program of Zhejiang Province~(2018C01004).

\bibliographystyle{named}
\bibliography{ijcai19}

\end{document}